%% file: main.tex
\documentclass[twoside]{article}

\usepackage{aistats2020}
\usepackage[round]{natbib}

\usepackage{hyperref}       
\usepackage{url}            
\usepackage{booktabs}       
\usepackage{nicefrac}       
\usepackage{microtype}      

\usepackage{latexsym,amssymb,amstext,amsfonts,amsmath,graphicx,setspace,mathtools,bm,color}
\usepackage{caption,subcaption}
\usepackage{comment}
\usepackage{gensymb}

\usepackage{enumerate} 
\usepackage{paralist}
\usepackage{float}
\usepackage{array}

\usepackage{multirow}

\usepackage{lipsum}

\newcommand{\sssection}[1]{\noindent\textbf{#1}\quad}

\begin{document}

%

%

\twocolumn[

\aistatstitle{Discovery and Separation of Features for\\Invariant Representation Learning}

\aistatsauthor{ Ayush Jaiswal, Rob Brekelmans, Daniel Moyer,\\ \textbf{Greg Ver Steeg, Wael AbdAlmageed, Premkumar Natarajan}\\ }

\aistatsaddress{ Information Sciences Institute, University of Southern California \\
\texttt{\{ajaiswal, gregv, wamageed, pnataraj\}@isi.edu, \{brekelma,moyerd\}@usc.edu} } ]

\begin{abstract}
  Supervised machine learning models often associate irrelevant nuisance factors with the prediction target, which hurts generalization. We propose a framework for training robust neural networks that induces invariance to nuisances through learning to discover and separate predictive and nuisance factors of data. We present an information theoretic formulation of our approach, from which we derive training objectives and its connections with previous methods. Empirical results on a wide array of datasets show that the proposed framework achieves state-of-the-art performance, without requiring nuisance annotations during training.
\end{abstract}

\input{sections/01_introduction.tex}

\input{sections/02_related_work.tex}

\input{sections/03_method.tex}

\input{sections/04_analysis.tex}

\input{sections/05_evaluation.tex}

\input{sections/06_conclusion.tex}

\bibliography{main}

\end{document}

%% file: sections/01_introduction.tex
\section{Introduction}
\label{sec:introduction}

Predictive models that incorporate irrelevant nuisance factors in the decision making process are vulnerable to poor generalization, especially when predictions need to be made on data with previously unseen configurations of nuisances~\citep{bib:infodropout,bib:achille2018,bib:vib,bib:uai}. This dependence on spurious connections between the prediction target and extraneous factors also makes models less reliable for practical use. Supervised machine learning models often learn such false associations due to the nature of the training objective and the optimization procedure~\citep{bib:uai}. Consequently, there is a growing interest in the development of training strategies that make supervised models robust through invariance to nuisances~\citep{bib:infodropout,bib:achille2018,bib:vib,bib:uai,bib:unifai,bib:vfae,bib:cvib,bib:iraf}.

Information theoretic and statistical measures have been traditionally used to prune out features that are unrelated to the prediction target~\citep{gao2016variational,bib:feat_select}. These approaches are not directly applicable to neural networks (NNs) that use complex raw data as inputs, e.g., images, speech signals, natural language text, etc. Motivated by the fact that NNs learn latent representations of data in the form of activations of their hidden layers, recent works~\citep{bib:infodropout,bib:achille2018,bib:vib,bib:uai} have framed robustness to nuisance for NNs as the task of invariant representation learning. In this formulation, latent representations of NNs are made invariant through (1) na\"ively training models with large variations of nuisance factors (e.g., through data augmentation)~\citep{bib:data_aug_1,bib:data_aug_2}, or (2) training mechanisms that lead to the exclusion of nuisance factors from the latent representation~\citep{bib:infodropout,bib:vib,bib:uai,bib:unifai,bib:vfae,bib:cvib,bib:cai}. Previous works have shown that the latter is a more effective approach for learning invariant representations~\citep{bib:infodropout,bib:vib,bib:uai,bib:vfae,bib:cvib,bib:ropad,bib:niesr}.

In this work, we propose a method for inducing nuisance-invariance through learning to discover and separate predictive and nuisance factors of data. The proposed framework generates a complete representation of data in the form of two independent embeddings --- one for encoding predictive factors and another for nuisances. This is achieved by augmenting the target-prediction objective, during training, with a reconstruction loss for decoding data from the said complete representation while simultaneously enforcing information constraints on the two constituent embeddings.

We present an information theoretic formulation of this approach and derive two equivalent training objectives as well as its relationship with previous methods. The proposed approach does not require annotations of nuisance factors for inducing their invariance, which is desired both in practice and in theory~\citep{bib:achille2018}. Extensive experimental evaluation shows that the proposed framework outperforms previous state-of-the-art methods that do not employ nuisance annotations as well as those that do.

%% file: sections/02_related_work.tex
\section{Related Work}
\label{sec:related_work}

Several approaches for invariant representation learning within neural networks have been proposed in recent works. These can be generally divided into two groups: those that require annotations~\citep{bib:nnmmd,bib:hcv,bib:vfae,bib:cvib,bib:cai,bib:lfr} for the undesired factors of data, and those that do not~\citep{bib:infodropout,bib:vib,bib:uai}. Methods that require annotations of undesired factors are suitable for targeted removal of specific kinds of information from the latent space. They are, thus, applicable for removing those factors of data that are correlated with the prediction target but are undesired due to external reasons, e.g., biased variables corresponding to race and gender. However, the need for annotations of undesired factors during training is a constraint that may not always be reasonable for every application, which limits the usage of these methods.

Numerous models have been introduced that use annotations of undesired factors during training. The NN+MMD model of \cite{bib:nnmmd} minimizes Maximum Mean Discrepancy (MMD)~\citep{bib:mmd} as a regularizer for removing undesired information. The Variational Fair Autoencoder (VFAE)~\citep{bib:vfae} is a Variational Autoencoder (VAE) that accounts for the unwanted factor in the probabilistic generative process and additionally uses MMD to boost invariance. \cite{bib:hcv} use the Hilbert-Schmidt Independence Criterion (HSIC)~\citep{bib:hsic} instead of MMD in their HSIC-constrained VAE (HCV) for the same purpose. The Controllable Adversarial Invariance (CAI) model~\citep{bib:cai} uses the gradient reversal trick~\citep{bib:dann} to penalize a model if it encodes the undesired factors. Fader Networks~\citep{bib:fn} when applied to the invariance task also reduce to the CAI model. \cite{bib:cvib} present a conditional form of the Information Bottleneck (IB) objective~\citep{bib:ib} and optimize its variational bound (CVIB).

Annotation-free invariance methods cannot be used to remove biased information from the latent embedding because there is no way to tell whether a predictive factor is biased or not. However, this class of methods is well suited for learning representations invariant to nuisances, largely due to two reasons --- (1) these approaches require no additional annotation besides the prediction target, making them more widely applicable in practice, and (2) it is known~\citep{bib:achille2018} that nuisance annotations are not necessary for learning minimal yet sufficient representations of data for a prediction task.

Training a supervised model with the Information Bottleneck (IB) objective~\citep{bib:ib} can compress out all nuisances from the latent embedding under optimality~\citep{bib:achille2018}. However, in practice, IB is very difficult to optimize~\citep{bib:infodropout,bib:vib}. Recent works have approximated IB variationally~\citep{bib:vib} or in the form of information dropout~\citep{bib:infodropout}. The Unsupervised Adversarial Invariance (UAI)~\citep{bib:uai} framework achieves invariance to nuisances by learning a split representation of data through competition between target-prediction and data-reconstruction objectives. The proposed framework builds upon IB but learns to represent both predictive and nuisance data factors similar to UAI. We show later in Section~\ref{sec:analysis} that the UAI model is a relaxation of the proposed framework. Furthermore, results in Section~\ref{sec:evaluation} show that the proposed model outperforms both an exact IB method and UAI.

%% file: sections/03_method.tex
\section{Separation of Predictive and Nuisance Factors of Data}
\label{sec:method}

The working mechanism of neural networks can be interpreted as the mapping of data samples $x$ to latent codes $z$ (activations of an intermediate hidden layer) followed by the inference of the target $y$ from $z$, i.e., the sequence $x \xrightarrow{} z \xrightarrow{} y$. The goal of this work is to learn $z$ that are maximally informative for predicting $y$ but are invariant to all nuisance factors $s$. Our approach for generating nuisance-free embeddings involves learning a complete representation of data in the form of two independent embeddings, $z_p$ and $z_n$, where $z_p$ encodes only those factors that are predictive of $y$ and $z_n$ encodes the nuisance information.

In order to learn $z_p$ and $z_n$, we augment the prediction objective $\mathbb{E} \log p(y| z_p)$ with a reconstruction objective $\mathbb{E} \log p(x | \{ z_p, z_n \})$ for decoding $x$ from the complete representation $\{z_p, z_n\}$ while imposing information constraints on $z_p$ and $z_n$ in the form of mutual information measures: $I(z_p : x)$, $I(z_n : x)$, and $I(z_p : z_n)$. The reconstruction objective and the information constraints encourage the learning of information-rich embeddings~\citep{bib:recon_2,bib:recon_1} and promote the separation of predictive factors from nuisances into the two embeddings. In the following sections, we present an information theoretic formulation of this approach and derive training objectives.

\subsection{Invariance to Nuisance through Information Discovery and Separation}

The Information Bottleneck (IB) method~\citep{bib:ib} aims to learn minimal representations of data that are sufficient for predicting $y$ from $x$~\citep{bib:infodropout}. The optimization objective of IB can be written as:
\begin{align}
    \max ~~&I(z:y) \label{eq:ib} \\
    ~~\text{s.t.} ~~&I(z:x) \leq I_c \nonumber
\end{align}
where the goal is to maximize the predictive capability of $z$ while constraining how much information $z$ can encode. The method intuitively brings about a trade-off between the prediction capability of $z$ and its information theoretic ``size'', also known as the channel capacity or rate. This is exactly the rate-distortion trade-off~\citep{bib:ib}. While optimizing this objective is, in theory, sufficient~\citep{bib:achille2018} for getting rid of nuisance factors with respect to $y$, the optimization is difficult and relies on a powerful encoder $z = \text{Encoder}(x)$ that is capable of disentangling factors of data effectively such that only nuisance information is compressed away. In practice, this is hard to achieve directly and methods that help the encoder better separate predictive factors from nuisances (ideally, at an atomic level~\citep{bib:uai}) are expected to perform better by retaining more predictive information while being invariant to nuisances.

Our approach for improving this separation of predictive and nuisance factors is to more explicitly learn a complete representation of data that comprises two independent embeddings: $z_p$ for encoding predictive factors and $z_n$ for nuisances. The proposed optimization objective can be written as:
\begin{align}
    \max ~~&\alpha I(z_p:y) + I(x:\{ z_p, z_n \}) \label{eq:multi_obj_1} \\
    ~~\text{s.t.} ~~&I(z_p:x) \leq I_c \nonumber \\
                    &I(z_p:z_n) = 0 \nonumber
\end{align}
where $\alpha$ determines the relative importance of the two mutual information terms. The optimization objective in Equation~\ref{eq:multi_obj_1} can be relaxed with multipliers such that the objective $J$ is:
\begin{align}
    J &= \alpha I(z_p:y) + I(x:\{ z_p, z_n \}) \nonumber \\
    &\qquad\qquad- \lambda I(z_p:x) - \gamma I(z_p:z_n)  \label{eq:multi_obj_1_relaxed_first}
\end{align}
where $\lambda$ and $\gamma$ denote multipliers for the $I(z_p:x)$ and $I(z_p:z_n)$ constraints, respectively. The optimization of $I(z_p:y)$ and $I(x:\{ z_p, z_n \})$ is straightforward through their variational bounds~\citep{bib:vib,bib:cvib}: $\mathbb{E} \log p(y| z_p)$ and $\mathbb{E} \log p(x | \{ z_p, z_n \})$, respectively. We discuss next the optimization of the $I(z_p:x)$ and $I(z_p:z_n)$ terms for inducing the desired information separation.

\subsection{Embedding Compression}
\label{subsec:echo}

We directly optimize the mutual information $I(z_p:x)$ in Equation~\ref{eq:multi_obj_1_relaxed_first} for restricting the flow of information from $x$ to $z_p$. In IB terminology, this is also referred to as the compression of the $z_p$ embedding. We compute a simple exact expression for this mutual information using the recently developed method of Echo noise~\citep{bib:echo}, which takes the same shift-and-scale form as a VAE, but replaces the standard Gaussian noise with an implicit sampling procedure. This is in direct contrast with previous IB methods for invariance~\citep{bib:infodropout,bib:vib}, which optimize bounds on the compression rate. The encoding $z_p$ is calculated as:
\begin{align}
    z_p = f_p(x) + S_p(x) \varepsilon_p \label{eq:reparam}
\end{align}
where $f_p$ and $S_p$ are parameterized by neural networks with bounded output activations and the noise $\varepsilon_p$ is calculated recursively using Equation~\ref{eq:reparam} on iid samples $x^{\ell}$ from the data distribution $q_{\text{data}}(x)$ as:
\begin{align}
    \varepsilon_p & = f_p(x^{(0)}) + S_p(x^{(0)})\Big( f_p(x^{(1)})  +  S_p(x^{(1)}) \big( ... 
\end{align}
Thus, the noise corresponds to an infinite sum that repeatedly applies Equation~\ref{eq:reparam} to additional input samples.  The key observation here is that the original training sample $x$ is also an iid sample from the input. By simply relabeling the sample indices $\ell$, we can see that the distributions of $z_p$ and $\varepsilon_p$ match in the limit. This yields~\citep{bib:echo} an exact, analytic form for the mutual information:
\begin{align}
    I(z_p:x) = - \mathbb{E}_{x} \log  |\det S_p(x)| \label{eq:echo_mi}
\end{align}
Intuitively, given that (1) $I(z_p:x) = H(z_p) - H(z_p|x)$ and that (2) the $\varepsilon_p$ and $z_p$ distributions match, the entropy in a conditional and unconditional draw from $z_p$ differ only by the scaling factor $S_p(x)$. We use Equation~\ref{eq:echo_mi} to calculate the $I(z_p:x)$ term in our objective.





\subsection{Independence between the Embeddings}

The exact form of $I(z_p:z_n)$ is much more difficult to minimize than the $I(z_p:x)$ term described in Section~\ref{subsec:echo}. We explore two approaches for enforcing independence between $z_p$ and $z_n$ --- (1) independence through compression of both $z_p$ and $z_n$, and (2) Hilbert-Schmidt Independence Criterion. We also present the corresponding complete training objectives.

\subsubsection*{Independence through Compression}

The objective in Equation \ref{eq:multi_obj_1_relaxed_first} can be re-arranged to contain only terms limiting the information in each embedding. We first state an identity based on the chain rule of mutual information:
\begin{flalign}
    I(z_p:z_n) = I(z_p:x) - I(z_p:x|z_n) + I(z_p:z_n|x) \label{eq:chain_rule_1}
    ~~~~~~\raisetag{1\normalbaselineskip}
\end{flalign}
We next inspect the $I(z_p:z_n|x)$ term in this identity following~\citep{bib:cvib}:
\begin{align}
    I(z_p:z_n|x) &= H(z_p|x) - H(z_p|x,z_n) \nonumber \\
    &= H(z_p|x) - H(z_p|x) ~~=~~ 0 \label{eq:zero_mi}
\end{align}
which is intuitively true because $z_p$ and $z_n$ are computed only from $x$. Thus, we get:
\begin{align}
    I(z_p:z_n) = I(z_p:x) - I(z_p:x|z_n) \label{eq:chain_rule_1_final}
\end{align}
This gives us an alternate way for computing $I(z_p:z_n)$ but the $I(z_p:x|z_n)$ is still difficult to calculate in this expression. In order to simplify this further, we use another key identity: 
\begin{align}
    I(z_p:x|z_n) = I(x:\{ z_p, z_n \}) - I(z_n:x) \label{eq:chain_rule_2}
\end{align}
Substituting for $I(z_p:x|z_n)$ in Equation~\ref{eq:chain_rule_1_final}, we get:
\begin{flalign}
    I(z_p:z_n) = I(z_p:x) + I(z_n:x) - I(x:\{ z_p, z_n \}) \label{eq:e1e2_final}
    ~~~~~~\raisetag{1\normalbaselineskip}
\end{flalign}
The expression for $I(z_p:z_n)$ in Equation~\ref{eq:e1e2_final} allows for the enforcement of independence between $z_p$ and $z_n$ by optimizing their joint and individual mutual information with $x$. Using this identity, Equation~\ref{eq:multi_obj_1_relaxed_first} simplifies into two ``relevant information'' terms and two compression terms as follows:
\begin{align}
    J &=   \alpha I(z_p:y) + I(x:\{ z_p, z_n \}) \nonumber \\
    &\qquad\qquad- \lambda I(z_p:x) - \gamma I(z_p:z_n) \nonumber  \\
    &=  \alpha I(z_p:y) + I(x:\{ z_p, z_n \})- \lambda I(z_p:x) \nonumber \\
    &\qquad\qquad- \gamma  I(z_p:x) - \gamma   I(z_n:x) + \gamma  I(x:\{ z_p, z_n \}) \nonumber \\
    & = \alpha I(z_p:y) + (1 + \gamma) I(x:\{ z_p, z_n \}) \nonumber \\
    &\qquad\qquad- (\lambda + \gamma) I(z_p:x) - \gamma I(z_n:x) \label{eq:multi_obj_1_relaxed}
\end{align}
Intuitively, this corresponds to maximizing $I(z_p:y)$ and $I(x:\{ z_p, z_n \})$ while compressing $z_p$ more than $z_n$. The multipliers $\lambda$ and $\gamma$ can be separately tuned to weight the compression of $z_p$ and $z_n$. We calculate each of the compression terms using the method described in Section~\ref{subsec:echo}. The final training objective after substituting for the compression losses and the variational bounds on $I(z_p:y)$ and $I(x:\{ z_p, z_n \})$ is termed as DSF-C and can be written as:
\begin{flalign}
    &\hat{J}_{\text{DSF-C}} = \alpha \mathbb{E} \log p(y| z_p) + (1 + \gamma) \mathbb{E} \log p(x | \{ z_p, z_n \}) \nonumber \\
    &~~+ (\lambda + \gamma) \mathbb{E} \log  |\det S_p(x)| + \gamma \mathbb{E} \log  |\det S_n(x)| 
    ~~~~~~~\raisetag{1\normalbaselineskip}\label{eq:training_obj_1}
\end{flalign}

\subsubsection*{Hilbert-Schmidt Independence Criterion}

Independence between $z_p$ and $z_n$ can also be achieved through the optimization of the Hilbert-Schmidt Independence Criterion (HSIC)~\citep{bib:hsic} between the two embeddings. HSIC is a kernel generalization of covariance and constructing a penalty out of the Hilbert-Schmidt operator norm  of a kernel covariance pushes variables to be mutually independent~\citep{bib:hcv}. This provides an intuitive and a more ``direct'' option to enforce the independence constraint on $z_p$ and $z_n$.

The HSIC estimator~\citep{bib:hcv} is defined for variables $u$ and $v$ with kernels $h$ and $k$ respectively. Assuming that the kernels are both universal, the following is an estimator of a two-component HSIC:
\begin{align}
    \text{HSIC}\bigl( \{(u,v)\}_{n}^N\bigr) &= \frac{1}{N^2}\sum_{i,j}k(u_i,u_j)h(v_i,v_j) \nonumber \\
    &~~~~+ \frac{1}{N^4} \sum_{i,j,k,l} k(u_i,u_j)h(v_k,v_l) \nonumber \\
    &~~~~+ \frac{2}{N^3} \sum_{i,j,k}k(u_i,u_j)h(v_i,v_k)
\end{align}
This is differentiable and can be used directly as a penalty on the ``dependent-ness'' of variables. The final training objective is termed as DSF-H (for short) and can be written as:
\begin{align}
    \hat{J}_{\text{DSF-H}} = \ &\alpha \mathbb{E} \log p(y| z_p) + \mathbb{E} \log p(x | \{ z_p, z_n \}) \nonumber \\
    &+ \lambda \mathbb{E} \log  |\det S_p(x)| - \gamma \text{HSIC}(z_p, z_n) \label{eq:training_obj_2}
\end{align}

\subsection{Model Implementation and Training}

We implemented the models in Keras with TensorFlow backend. We used the Adam optimizer with $10^{-4}$ learning rate and $10^{-4}$ weight decay. The multiplier $\alpha$ was fixed at $1$; $\lambda$ and $\gamma$ were tuned through grid-search over $\{10^{-4}, 10^{-3}, 10^{-2}, 10^{-1}\}$. We used diagonal $S(x)$ with number of samples limited to the batch-size instead of the infinite sum as described in Section~\ref{subsec:echo}. 

%% file: sections/04_analysis.tex
\section{Analysis}
\label{sec:analysis}

In this section we derive a relationship between the proposed framework and the UAI model~\citep{bib:uai}. This analysis is useful in understanding both the proposed model and UAI. We show that the UAI objective is a relaxation of the objective we propose in Equation~\ref{eq:multi_obj_1_relaxed_first}, which additionally demonstrates the superiority of the proposed framework for learning nuisance-invariant representations.

{
\def \fs {0.45}
\def \sfs {0.9}
\begin{figure*}
\centering
\begin{subfigure}{\fs\textwidth}
\centering
\includegraphics[width=\sfs\textwidth]{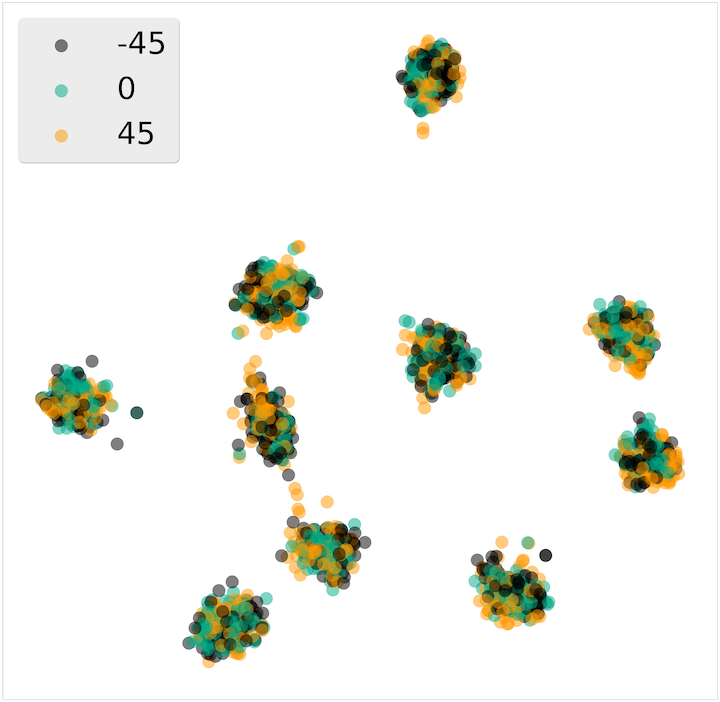}
\caption*{$z_p$ embedding}
\end{subfigure}
\begin{subfigure}{\fs\textwidth}
\centering
\includegraphics[width=\sfs\textwidth]{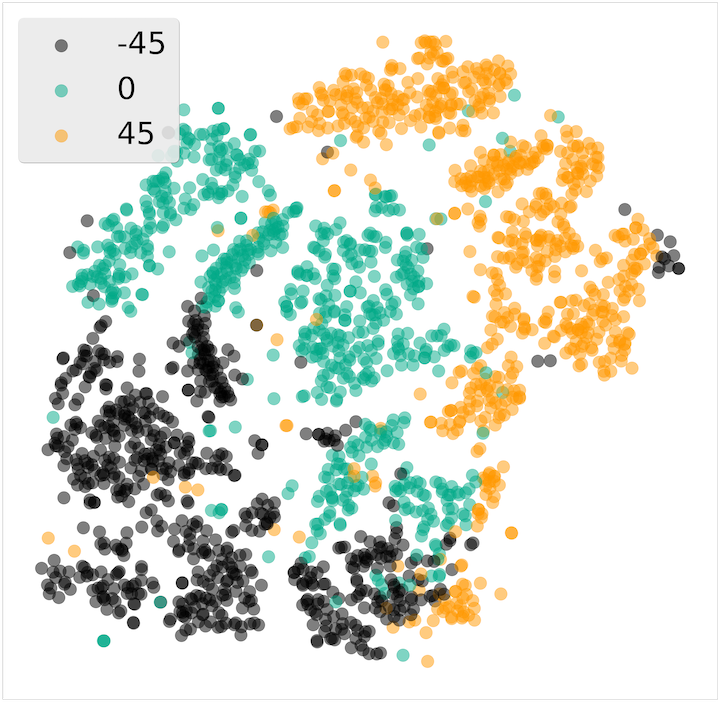}
\caption*{$z_n$ embedding}
\end{subfigure}
\caption{t-SNE visualization of $z_p$ and $z_n$ embeddings of MNIST-ROT images labeled by rotation angle. As desired, the $z_p$ embedding does not encode rotation information, which migrates to $z_n$.}
\label{fig:tsne_mnist}
\end{figure*}
}

We start with rearranging the proposed objective in Equation~\ref{eq:multi_obj_1_relaxed_first} using the identity from Equation~\ref{eq:chain_rule_1_final} as $I(z_p:x) = I(z_p:z_n) + I(z_p:x|z_n)$. This gives us:
\begin{align}
 J &=   \alpha I(z_p:y) + I(x:\{ z_p, z_n \}) \nonumber \\
 &\qquad\qquad- \lambda I(z_p:x) - \gamma I(z_p:z_n)  \nonumber \\
  &=   \alpha I(z_p:y) + I(x:\{ z_p, z_n \}) - \lambda I(z_p:z_n) \nonumber \\
  &\qquad\qquad- \lambda I(z_p:x|z_n) - \gamma I(z_p:z_n) \nonumber  \\
   &=   \alpha I(z_p:y) + \Bigl \{ I(x:\{ z_p, z_n \}) - \lambda I(z_p:x|z_n) \Bigr \} \nonumber \\
   &\qquad\qquad- (\lambda + \gamma) I(z_p:z_n) \label{eq:pre_uai}
\end{align}
Recall that the chain rule for mutual information in Equation~\ref{eq:chain_rule_2} implies that $I(x:\{ z_p, z_n \}) =  I(z_n:x) + I(z_p:x | z_n)$. In the expression $I(x:\{ z_p, z_n \}) - \lambda I(z_p:x | z_n)$ in braces above, our objective simply downweights the $I(z_p:x | z_n)$ component of $I(x:\{ z_p, z_n \})$ as:
\begin{align}
    &I(x:\{ z_p, z_n \}) - \lambda I(z_p:x | z_n) \nonumber \\
    &\qquad\qquad= I(z_n:x) + (1 - \lambda) I(z_p:x | z_n) \label{eq:lambda_info}
\end{align}
This is equivalent to calculating $I(x:\{ \tilde{z}_p, z_n \})$ for a noisified $\tilde{z}_p = \psi(z_p)$ such that:
\begin{align}
    I(x:\{ \tilde{z}_p, z_n \}) &= I(z_n:x) + I(\tilde{z}_p:x | z_n) \\
    I(\tilde{z}_p:x | z_n) &= (1 - \lambda) I(z_p:x | z_n) \label{eq:noisy_diff}
\end{align}
where the free parameter $\lambda$ can be chosen to enforce this relationship. In particular, $\lambda$ depends on $I(z_p:x|\{ \tilde{z_p}, z_n\})$, which measures the information about $x$ that is destroyed by adding noise through $\psi$. We derive this result using the chain rule for mutual information in two different ways:
\begin{flalign}
    I(x:\{ \tilde{z}_p, z_p, z_n \}) = I(x:\{ \tilde{z}_p, z_n \} ) + I(z_p:x| \{ \tilde{z}_p, z_n\})& ~~\raisetag{1\normalbaselineskip} \nonumber \\
    =  I(x:\{ z_p, z_n\}) + I(\tilde{z}_p:x| \{z_p , z_n\})& ~~\raisetag{1\normalbaselineskip} \nonumber \\
    \therefore~ I(x:\{ z_p, z_n \} ) = I(x:\{ \tilde{z}_p , z_n\}) + I(z_p:x| \{ \tilde{z}_p, z_n\})& \label{eq:mi_gap}
\end{flalign}
In Equation~\ref{eq:mi_gap}, we used the fact that $I(\tilde{z}_p:x| \{z_p , z_n\}) = 0$ by the data processing inequality. Using the chain rule again, we know that both $I(x:\{ z_p, z_n\})$ and $I(x:\{ \tilde{z}_p, z_n\})$ contain a $I(z_n:x)$ term. Canceling out $I(z_n:x)$ and rearranging to match the form of Equation \ref{eq:noisy_diff}, we obtain:
\begin{align}
    I(\tilde{z}_p:x | z_n) &= I(z_p:x | z_n ) - I(z_p:x| \{ \tilde{z}_p, z_n\})
\end{align}
Thus, the information using a noisy $\tilde{z}_p$ instead of $z_p$ differs by a term of $I(z_p:x| \{ \tilde{z}_p, z_n\})$. Since $\lambda$ is a free parameter and does not appear elsewhere in the objective, it can be set to satisfy Equation~\ref{eq:noisy_diff}. The objective function in Equation~\ref{eq:pre_uai} can be rewritten with this noisy $\tilde{z}_p$ as:
\begin{align}
    J = \alpha I(z_p:y) + I(x:\{ \tilde{z}_p, z_n \}) - \gamma I(z_p:z_n) \label{eq:original_uai}
\end{align}
Although not explored in~\citep{bib:uai}, Equation~\ref{eq:original_uai} describes an information theoretic formulation of UAI.

\begin{table*}
\centering
\caption{\label{tab:mnist_rot}MNIST-ROT results presented as accuracy for $\Theta = \{0, \pm 22.5, \pm 45\}$ that the models were trained on and unseen $\pm 55$ and $\pm 65$ angles. VFAE could not be evaluated for $\pm 55$ and $\pm 65$ as it uses $s$ as input for encoding $x$ and cannot be used for unseen $s$. The $y$-accuracy should be high but $s$-accuracy should be close to random chance (0.20). RI indicates relative improvement in error-rate between DSF-H (ours) and UAI (previous best).}
\setlength{\tabcolsep}{0.725em} 
\begin{tabular}{c c c c c c c c c c}
 \cmidrule(r){1-2}\cmidrule(lr){3-6} \cmidrule(l){7-10}
 \textbf{Acc.} & $\boldsymbol{\theta}$ & \textbf{VFAE} & \textbf{CAI} & \textbf{CVIB} & \textbf{UAI} & \textbf{DSF-E} & \textbf{DSF-C} & \textbf{DSF-H} & \textbf{RI} \\
\cmidrule(r){1-1} \cmidrule(lr){2-2}  \cmidrule(lr){3-3}  \cmidrule(lr){4-4}  \cmidrule(lr){5-5}  \cmidrule(lr){6-6} \cmidrule(lr){7-7} \cmidrule(lr){8-8} \cmidrule(l){9-10}
\multirow{3}{*}{$y$} & $\Theta$ & 0.953 & 0.958 & 0.960 & 0.977 & 0.980 & \textbf{0.981 $\pm$ 0.001} & \textbf{0.981 $\pm$ 0.001} & 17\% \\
 & $\pm$55 & $\times$ & 0.826 & 0.819 & 0.856 & 0.865 & 0.869 $\pm$ 0.001 & \textbf{0.873 $\pm$ 0.002} & 12\% \\
 & $\pm$65 & $\times$ & 0.662 & 0.674 & 0.696 & 0.707 & 0.724 $\pm$ 0.002 & \textbf{0.732 $\pm$ 0.001} & 12\% \\
\cmidrule(r){1-1}
$s$ & $\Theta$ & 0.389 & 0.384 & 0.382 & 0.338 & \textbf{0.200} & \textbf{0.200 $\pm$ 0.001} & \textbf{0.200 $\pm$ 0.000} & 100\% \\
\cmidrule(r){1-2} \cmidrule(lr){3-6} \cmidrule(l){7-10}
\end{tabular}
\end{table*}

\begin{table*}
\centering
\caption{\label{tab:mnist_dil}MNIST-DIL results presented as accuracy for various kernel sizes $\kappa$ (positive for dilation and negative for erosion). Models were trained on MNIST-ROT but tested on MNIST-DIL. RI indicates relative improvement in error-rate between DSF-H (ours) and UAI (previous best).}
\setlength{\tabcolsep}{0.8em} 
\begin{tabular}{c c c c c c c c c c}
 \cmidrule(r){1-2} \cmidrule(lr){3-6} \cmidrule(l){7-10}
 \textbf{Acc.} & $\boldsymbol{\kappa}$ & \textbf{VFAE} & \textbf{CAI} & \textbf{CVIB} & \textbf{UAI} & \textbf{DSF-E} & \textbf{DSF-C} & \textbf{DSF-H} & \textbf{RI} \\
\cmidrule(r){1-1} \cmidrule(lr){2-2}  \cmidrule(lr){3-3}  \cmidrule(lr){4-4}  \cmidrule(lr){5-5}  \cmidrule(lr){6-6} \cmidrule(lr){7-7} \cmidrule(lr){8-8} \cmidrule(l){9-10}
\multirow{4}{*}{$y$} & -2 & 0.807 & 0.816 & 0.844 & 0.880 & 0.891 & 0.899 $\pm$ 0.002 & \textbf{0.907 $\pm$ 0.001} & 22\% \\
 & 2 & 0.916 & 0.933 & 0.933 & 0.958 & 0.964 & 0.966 $\pm$ 0.001 & \textbf{0.970 $\pm$ 0.001} & 28\% \\
 & 3 & 0.818 & 0.795 & 0.846 & 0.874 & 0.887 & 0.889 $\pm$ 0.002 & \textbf{0.892 $\pm$ 0.002} & 14\% \\
 & 4 & 0.548 & 0.519 & 0.586 & 0.606 & 0.608 & 0.608 $\pm$ 0.002 & \textbf{0.610 $\pm$ 0.002} & 1\% \\
\cmidrule(r){1-2} \cmidrule(lr){3-6} \cmidrule(l){7-10}
\end{tabular}
\end{table*}

The UAI model uses independent multiplicative Bernoulli noise to create a noisy $\tilde{z}_p = \psi(z_p)$, which is then used alongside $z_n$ in a variational decoder maximizing $I(x: \{\tilde{z}_p, z_n\})$. This has the indirect effect of regularizing $I(z_p:x)$, since nuisance information cannot be reliably passed through $\tilde{z}_p$ for the reconstruction task. In contrast, the proposed objective in Equation~\ref{eq:multi_obj_1_relaxed_first} directly constrains the information channel between $x$ and $z_p$, which guarantees invariance to nuisance at optimality~\citep{bib:achille2018}.

%% file: sections/05_evaluation.tex
\section{Experimental Evaluation}
\label{sec:evaluation}

Empirical results are presented on four datasets --- MNIST-ROT~\citep{bib:uai}, MNIST-DIL~\citep{bib:uai}, Multi-PIE~\citep{bib:mpie}, and Chairs~\citep{bib:chairs}, following previous works~\citep{bib:nnmmd,bib:vfae,bib:cvib,bib:cai,bib:uai}. Samples of images in these datasets are shown in the supplementary material. The proposed model is compared with previous state-of-the-art: VFAE, CAI, CVIB, and UAI. Results are also reported for an ablation version of our framework, DSF-E, which models the IB objective in Equation~\ref{eq:ib}. We optimize an exact expression for $I(z_p:x)$, as presented in Section~\ref{subsec:echo}, for DSF-E. Hence, we do not evaluate methods that are similar to the ablation model but indirectly optimize $I(z_p:x)$~\citep{bib:infodropout,bib:vib}.

The accuracy of predicting $y$ from $z_p$ is reported for the trained models. Additionally, the accuracy of predicting $s$ is reported as a measure of invariance using two layer neural networks that were trained \textit{post hoc} to predict known $s$ from $z_p$, following previous works~\citep{bib:uai,bib:cvib}. While the $y$-accuracy is desired to be high, the $s$-accuracy should be close to random chance of $s$ for true invariance. Results of the full version of our model are reported as mean and standard-deviation based on five runs. We also report relative improvements (\%) in error-rate over previous best models. The error-rate for $s$ is calculated as the difference between $s$-accuracy and random chance. Furthermore, t-SNE~\citep{bib:tsne} visualization of the $z_p$ and $z_n$ embeddings are presented for the DSF-H version of the proposed model for visualizing the separation of nuisance factors.


{
\def \fs {0.45}
\def \sfs {0.89}
\begin{figure*}
\centering
\captionsetup[subfigure]{aboveskip=1pt,belowskip=2pt}
\begin{subfigure}{\fs\textwidth}
\centering
\includegraphics[width=\sfs\textwidth]{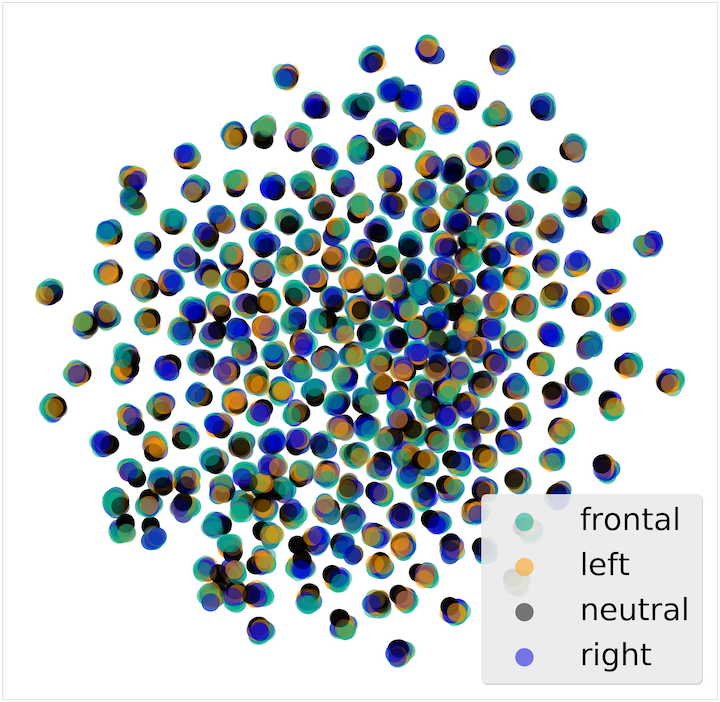}
\caption*{$z_p$ -- labeled by illumination}
\end{subfigure}
\begin{subfigure}{\fs\textwidth}
\centering
\includegraphics[width=\sfs\textwidth]{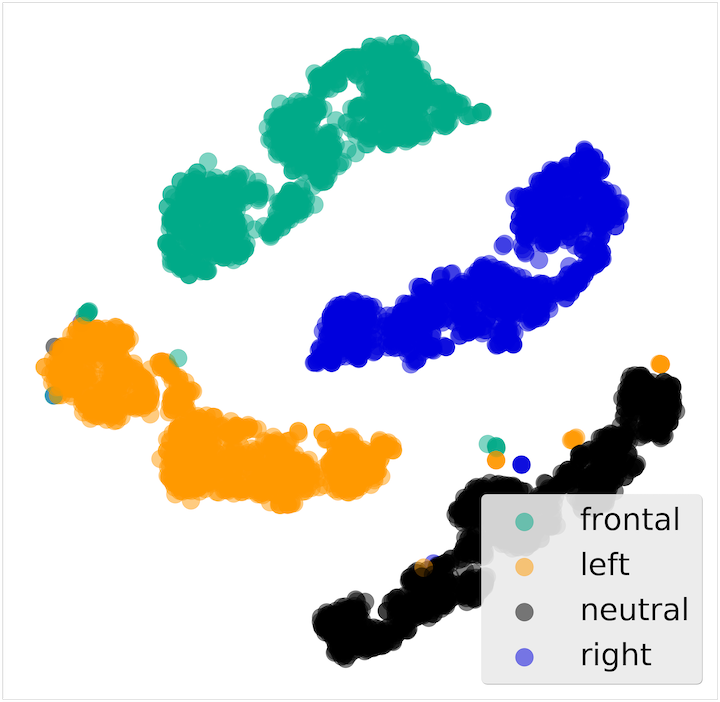}
\caption*{$z_n$ -- labeled by illumination}
\end{subfigure}
\begin{subfigure}{\fs\textwidth}
\centering
\includegraphics[width=\sfs\textwidth]{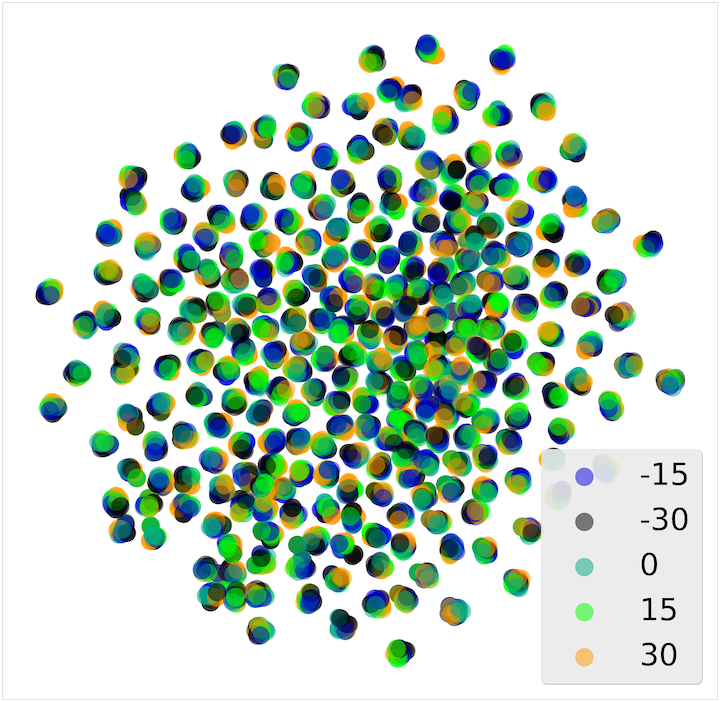}
\caption*{$z_p$ -- labeled by pose}
\end{subfigure}
\begin{subfigure}{\fs\textwidth}
\centering
\includegraphics[width=\sfs\textwidth]{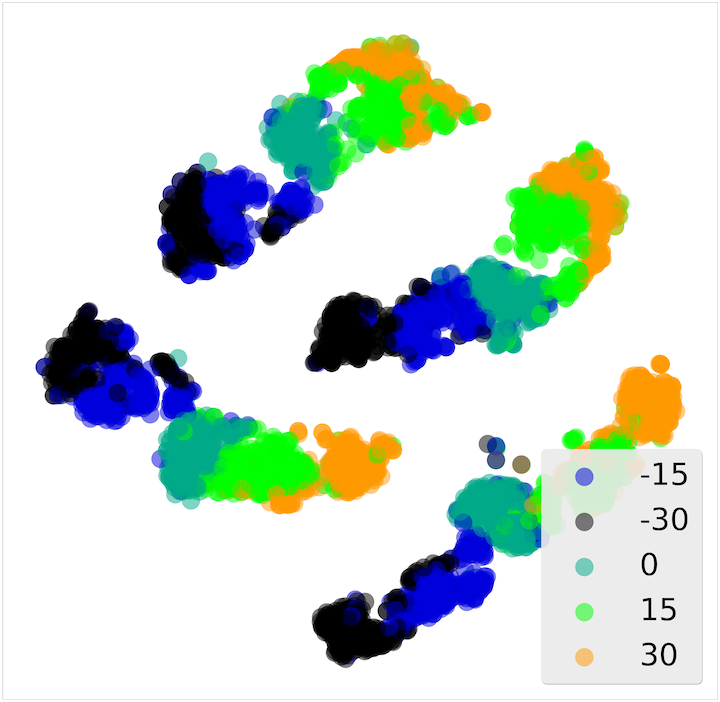}
\caption*{$z_n$ -- labeled by pose}
\end{subfigure}
\caption{t-SNE visualization of $z_p$ and $z_n$ embeddings of Multi-PIE images labeled by illumination (top row) and pose (bottom row). As desired, $z_p$ does not encode illumination and pose, both of which migrate to $z_n$.}
\label{fig:tsne_mpie}
\end{figure*}
}

\begin{table*}
\centering
\caption{\label{tab:mpie}Multi-PIE --- $y$-accuracy should be high but $s$-accuracy should be random chance (illumination (i): 0.25, pose (p): 0.20). Separate models were trained for illumination and pose for previous supervised invariance methods: VFAE, CAI, and CVIB. RI indicates relative improvement in error-rate over previous best (UAI).}
\setlength{\tabcolsep}{0.69em} 
\begin{tabular}{c c c c c c c c c c c c}
 \cmidrule(r){1-1} \cmidrule(lr){2-8} \cmidrule(l){9-12}
 \textbf{Acc.} & \multicolumn{2}{c}{\textbf{VFAE}} & \multicolumn{2}{c}{\textbf{CAI}} & \multicolumn{2}{c}{\textbf{CVIB}} & \textbf{UAI} & \textbf{DSF-E} & \textbf{DSF-C} & \textbf{DSF-H} & \textbf{RI} \\
\cmidrule(r){1-1} \cmidrule(lr){2-3} \cmidrule(lr){4-5}  \cmidrule(lr){6-7} \cmidrule(lr){8-8} \cmidrule(lr){9-9} \cmidrule(lr){10-10} \cmidrule(l){11-12}
 & $s$: i & $s$: p & $s$: i & $s$: p & $s$: i & $s$: p & & & & & \\
\cmidrule(lr){2-2}  \cmidrule(lr){3-3}  \cmidrule(lr){4-4}  \cmidrule(lr){5-5}  \cmidrule(lr){6-6} \cmidrule(lr){7-7}
$y$ & 0.67 & 0.62 & 0.76 & 0.77 & 0.51 & 0.46 & 0.82 & 0.83 & 0.85 $\pm$ 0.00 & \textbf{0.87 $\pm$ 0.01} & 28\% \\
$s$: i & 0.41 & 0.80 & 0.99 & 1.00 & 0.44 & 0.63 & 0.61 & \textbf{0.25} & \textbf{0.25 $\pm$ 0.02} & \textbf{0.25 $\pm$ 0.00} & 100\% \\
$s$: p & 0.65 & 0.29 & 0.98 & 0.98 & 0.45 & 0.28 & 0.32 & \textbf{0.20} & \textbf{0.20 $\pm$ 0.01} & \textbf{0.20 $\pm$ 0.00} & 100\% \\
\cmidrule(r){1-1} \cmidrule(lr){2-8} \cmidrule(l){9-12}
\end{tabular}
\end{table*}

\begin{table*}
\centering
\caption{\label{tab:chairs}Chairs results --- $y$-accuracy should be high but $s$-accuracy should be random chance (0.25). RI indicates relative improvement in error-rate between DSF-H (ours) and UAI (previous best).}
\setlength{\tabcolsep}{1.0em} 
\begin{tabular}{c c c c c c c c c}
 \cmidrule(r){1-1} \cmidrule(lr){2-5} \cmidrule(l){6-9}
 \textbf{Acc.} & \textbf{VFAE} & \textbf{CAI} & \textbf{CVIB} & \textbf{UAI} & \textbf{DSF-E} & \textbf{DSF-C} & \textbf{DSF-H} & \textbf{RI} \\
\cmidrule(r){1-1} \cmidrule(lr){2-2}  \cmidrule(lr){3-3}  \cmidrule(lr){4-4}  \cmidrule(lr){5-5}  \cmidrule(lr){6-6} \cmidrule(lr){7-7} \cmidrule(l){8-9}
$y$ & 0.72 & 0.68 & 0.67 & 0.74 & 0.84 & 0.86 $\pm$ 0.01 & \textbf{0.88 $\pm$ 0.01} & 54\% \\
$s$ & 0.37 & 0.69 & 0.52 & 0.34 & \textbf{0.25} & \textbf{0.25 $\pm$ 0.00} & \textbf{0.25 $\pm$ 0.00} & 100\% \\
\cmidrule(r){1-1} \cmidrule(lr){2-5} \cmidrule(l){6-9}
\end{tabular}
\end{table*}

{
\def \fs {0.45}
\def \sfs {0.9}
\begin{figure*}
\centering
\begin{subfigure}{\fs\textwidth}
\centering
\includegraphics[width=\sfs\textwidth]{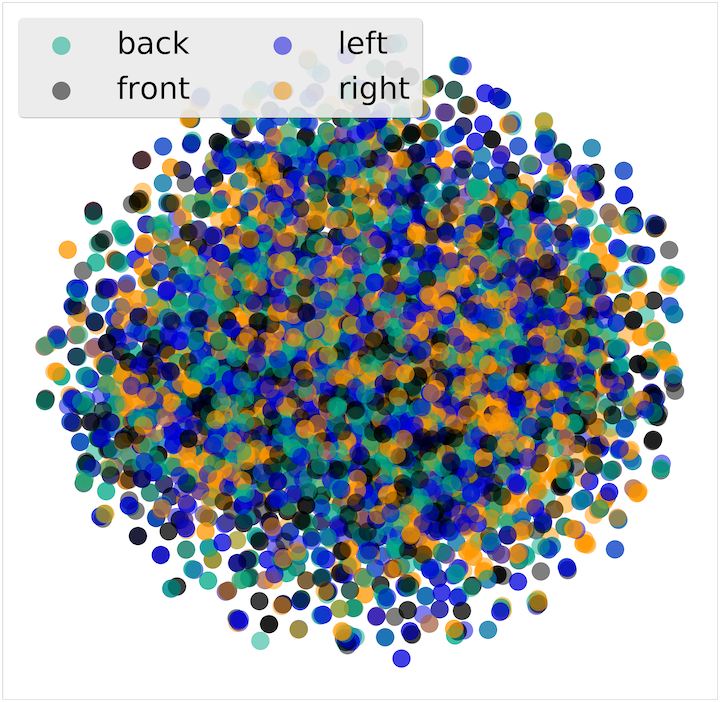}
\caption*{$z_p$ embedding}
\end{subfigure}
\begin{subfigure}{\fs\textwidth}
\centering
\includegraphics[width=\sfs\textwidth]{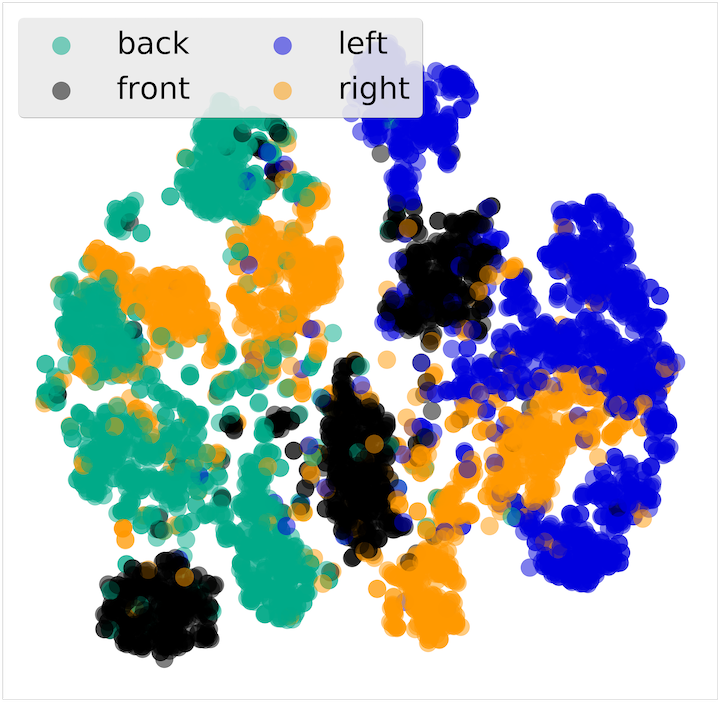}
\caption*{$z_n$ embedding}
\end{subfigure}
\caption{t-SNE visualization of $z_p$ and $z_n$ embeddings of Chairs images labeled by yaw orientation. As desired, the $z_p$ embedding does not encode orientation information, which migrates to $z_n$.}
\label{fig:tsne_chairs}
\end{figure*}
}

\sssection{MNIST-ROT:}This dataset was introduced in~\citep{bib:uai} as an augmented version of the MNIST~\citep{bib:mnist} dataset that contains digits rotated at angles $\theta \in \Theta = \{0, \pm 22.5, \pm 45\}$ for training. Evaluation is performed on digits rotated at $\theta \in \Theta$ as well as for $\pm 55$ and $\pm 65$. We use the same version of the dataset as~\citep{bib:uai}. The NN instantiation of the proposed model also follows UAI with two layers for encoding $x$ into $z_p$ and $z_n$, two layers for inferring $y$ from $z_p$, and three layers for reconstructing $x$ from $\{z_p, z_n\}$. The digit class is treated as $y$ and categorical $\theta$ as $s$. Table~\ref{tab:mnist_rot} presents results of our dataset, showing that all versions of the proposed framework outperform previous state-of-the-art models with DSF-H achieving the best $y$-accuracies. Furthermore, without using $s$-labels, all versions of the proposed framework achieve random chance $s$-accuracy ($0.20$), which indicates perfect invariance to rotation angle. Figure~\ref{fig:tsne_mnist} shows the t-SNE visualization of $z_p$ and $z_n$. As evident, $z_p$ does not cluster by rotation angle but $z_n$ does, which validates that this nuisance factor is separated out and encoded in $z_n$ instead of $z_p$.

\sssection{MNIST-DIL:}This variant of MNIST contains digits eroded or dilated with various kernel sizes $\kappa \in \{-2, 2, 3, 4\}$, as introduced in~\citep{bib:uai}. MNIST-DIL is used for further evaluating models trained on the MNIST-ROT dataset for varying stroke-widths, which is not explicitly controlled in MNIST-ROT but is implicitly present. Results in Table~\ref{tab:mnist_dil} show that all versions of the proposed framework outperform previous works by retaining more predictive information in $z_p$.

\sssection{Multi-PIE:}This is a dataset of face images of 337 subjects captured at 15 poses and 19 illumination conditions with various facial expressions. A subset of the data is prepared for this experiment that contains 264 subjects with images captured at five pose angles $\{0, \pm 15, \pm 30\}$ and four illumination conditions: neutral, frontal, left, and right. The subject identity is treated as $y$ while pose and illumination are treated as nuisances. The NN instantiation of the proposed model uses one layer each for encoding $z_p$ and $z_n$, and for predicting $y$ from $z_p$, while two layers are used for reconstructing $x$ from $\{z_p, z_n\}$. Table~\ref{tab:mpie} presents the results of this experiment, showing that all versions of the proposed model outperform previous works on both $y$-accuracy and $s$-accuracy of both illumination and pose. The t-SNE visualization of $z_p$ and $z_n$ in Figure~\ref{fig:tsne_mpie} shows that both illumination and pose information are separated out and encoded in $z_n$ instead of $z_p$, resulting in an invariant $z_p$ embedding.


\sssection{Chairs:}This dataset contains images of chairs at various yaw angles, which are binned into four orientations: front, back, left, and right. We use the same version of this dataset as~\citep{bib:uai} where the yaw angles do not overlap between the train and test sets. The NN instantiation follows~\citep{bib:uai} with two layers each for encoding $z_p$ and $z_n$ from $x$, predicting $y$ from $z_p$, and reconstructing $x$ from $\{z_p, z_n\}$. Results are summarized in Table~\ref{tab:chairs}, showing that all versions of the proposed framework outperform previous methods by a large margin on both $y$-accuracy and $s$-accuracy. All versions of the proposed framework also achieve random chance $s$-accuracy ($0.25$), which indicates perfect invariance to orientation, without using $s$-labels during training. Figure~\ref{fig:tsne_chairs} shows the t-SNE visualization of $z_p$ and $z_n$, further validating that the orientation information is separated out of $z_p$ and encoded in $z_n$.

%% file: sections/06_conclusion.tex
\section{Conclusion}
\label{sec:conclusion}

We have presented a novel framework for inducing nuisance-invariant representations in supervised NNs through learning to encode all information about the data while separating out predictive and nuisance factors into independent embeddings. We provided an information theoretic formulation of the approach and derived training objectives from it. Furthermore, we provided a theoretical analysis of the proposed model and derived a connection with the UAI model, showing that the proposed framework is superior to UAI. Empirical results on benchmark datasets show that the proposed framework outperforms previous works with large relative improvements.